\pdfoutput=1
\documentclass[10pt,twocolumn,letterpaper]{article}

\usepackage[pagenumbers]{cvpr}

\usepackage[accsupp]{axessibility}
\usepackage{graphicx}
\usepackage{amsmath}
\usepackage{amssymb}
\usepackage{booktabs, wrapfig}
\usepackage{multirow}
\usepackage{comment}
\usepackage{soul}

\usepackage[pagebackref,breaklinks,colorlinks]{hyperref}

\usepackage[capitalize]{cleveref}
\crefname{section}{Sec.}{Secs.}
\Crefname{section}{Section}{Sections}
\Crefname{table}{Table}{Tables}
\crefname{table}{Tab.}{Tabs.}

\def\cvprPaperID{7390}
\def\confName{CVPR}
\def\confYear{2024}

\begin{document}
\makeatletter

\makeatother

\newcommand{\cm}{\checkmark}

\newcommand{\newterm}[1]{\emph{#1}}
\newcommand{\TODO}[1]{$<$\textcolor{red}{#1}$>$}
\newcommand{\REV}[1]{\textcolor{blue}{#1}}
\newcommand{\REVNO}[1]{\st{#1}}
\newcommand{\final}[1]{#1}
\newcommand{\migue}[1]{[Migue]\textcolor{red}{#1}}
\newcommand{\emma}[1]{[Emma]\textcolor{blue}{#1}}
\newcommand{\viv}[1]{[Viv]\textcolor{cyan}{#1}}

\title{DUDF: Differentiable Unsigned Distance Fields with Hyperbolic Scaling}

\author{Miguel Fainstein$^{\text{1},\text{2}}$\\
{\tt\small miguelon.f98@gmail.com}
\and
Viviana Siless$^{\text{1}}$\\
{\tt\small viviana.siless@utdt.edu}
\and
Emmanuel Iarussi$^{\text{1},\text{3}}$\\
{\tt\small emmanuel.iarussi@utdt.edu}
\and
$^{\text{1}}$Universidad Torcuato Di Tella, $^{\text{2}}$FCEyN Universidad de Buenos Aires, $^{\text{3}}$CONICET
} 
\maketitle

\begin{abstract}

In recent years, there has been a growing interest in training Neural Networks to approximate Unsigned Distance Fields (UDFs) for representing open surfaces in the context of 3D reconstruction.
However, UDFs are non-differentiable at the zero level set which leads to significant errors in distances and gradients, generally resulting in fragmented and discontinuous surfaces. 
In this paper, we propose to learn a hyperbolic scaling of the unsigned distance field, which defines a new Eikonal problem with distinct boundary conditions.
This allows our formulation to integrate seamlessly with state-of-the-art continuously differentiable implicit neural representation networks, largely applied in the literature to represent signed distance fields.
Our approach not only addresses the challenge of open surface representation but also demonstrates significant improvement in reconstruction quality and training performance. 
Moreover, the unlocked field's differentiability allows the accurate computation of essential topological properties such as normal directions and curvatures, pervasive in downstream tasks such as rendering.
Through extensive experiments, we validate our approach across various data sets and against competitive baselines.
The results demonstrate enhanced accuracy and up to an order of magnitude increase in speed compared to previous methods.

\end{abstract}

\section{Introduction}
\label{sec:intro}
Surface representation is a fundamental aspect in the field of 3D geometry processing, with explicit methods such as meshes, point clouds, and voxelized representations being traditional choices.
Implicit surface representations, on the other hand, have been an integral part of the graphics pipeline for many decades. 
They encapsulate surfaces as the zero-level set of a function, providing a compact and continuous geometry representation.
The novelty in recent years has emerged from parameterizing these implicit functions with Neural Networks (NNs), combining their learning capabilities with the advantages of implicit representations.

\begin{figure}[t]
  \centering
  \includegraphics[width=1.0\linewidth]{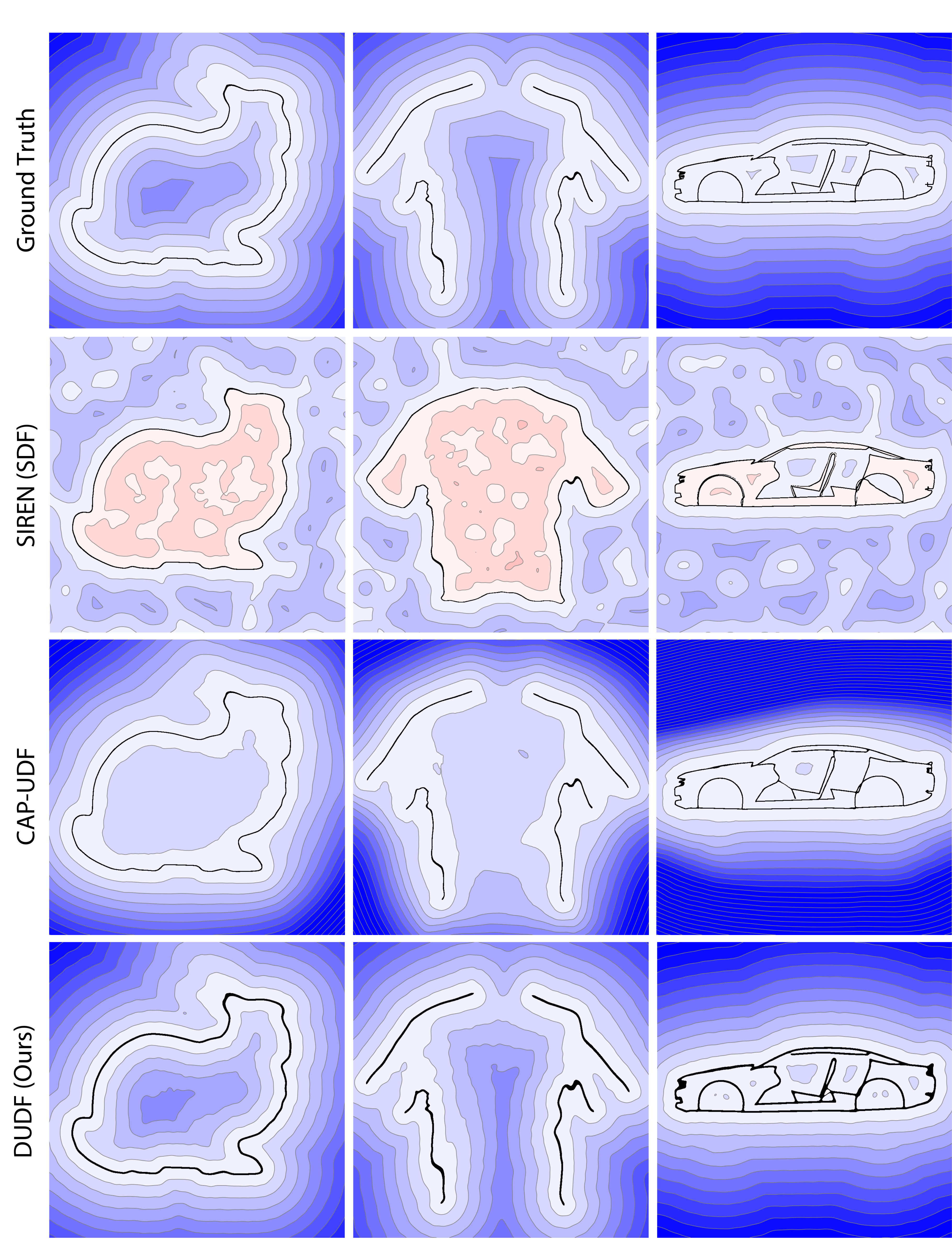}
   \caption{Distance field cross-sections comparative. 
    First row displays ground truth unsigned distance fields. 
    Methods based on signed distances, like SIREN  \cite{sitzmann2020implicit}, mistakenly fill gaps in open surfaces. 
    While methods based on unsigned distances such as CAP-UDF \cite{zhou2022learning} can represent non-watertight surfaces, the learned distance field do not approximate the true function.
   DUDF's differentiable formulation successfully overcomes these challenges.}
   \label{fig:teaser}
\end{figure}

Signed Distance Functions (SDFs) have traditionally been the chosen formulation for implicit surface representation \cite{curless1996volumetric,newcombe2011kinectfusion} due to their well-defined gradients and the ease they offer for computing constructive solid geometry operations and mesh reconstruction.
However, they are inherently limited to closed surfaces (Fig. \ref{fig:teaser}), which poses a significant challenge for representing open surfaces with implicit methods. 
This limitation arises from the inside/outside sign flip on the zero level set, which is impossible to define for surfaces that do not enclose a volume. 

The advent of Unsigned Distance Functions (UDFs) extended representation capabilities to open surfaces. 
However, this advancement introduced a challenge: the non-differentiable nature of the distance functions at the zero level set. 
Such non-differentiability leads to inaccuracies in distances and gradients learned by NNs, particularly near the surface where precision is paramount. 
This paper tackles the challenge of representing open surfaces with neural networks, focusing on a formulation that ensures the surface continuity and smoothness.
Our framework (DUDF) combines a hyperbolic scaling of the distance field, together with a new \textit{Eikonal} problem  featuring tailored boundary conditions. 
This formulation facilitates the training of continuously differentiable implicit neural representation networks, while preserving the essential attributes of the unsigned distance fields. 
Additionally, the differentiable nature of our representations enables precise calculations of crucial topological characteristics, including curvatures and normals, which are pervasive in downstream tasks such as rendering.
In contrast, earlier techniques depended on mesh reconstruction methods like Marching Cubes \cite{lorensen1998marching, guillard2022meshudf, zhou2022cap-udf} for rendering purposes.

We evaluate our framework with experiments that rigorously assess performance on a variety of challenging data sets and compare to competitive benchmarks in the field. 
Our method overcomes several limitations of previous approaches that struggled with the non-differentiability of UDFs.
Results demonstrate that our approach not only enhances the reconstruction quality of open surfaces but also accelerates the training process.
We believe the dual improvement in precision and efficiency enhance the applicability of neural UDFs representations in geometric processing for real-world scenarios.

\section{Related Works}
\label{sec:related_work}
The representation and reconstruction of three-dimensional geometry is a fundamental challenge in the field of computer graphics, vision, and computational geometry. We now briefly discuss the key contributions and methodologies adopted in recent literature to address the main challenges in the field, distinguishing between closed and open surface paradigms.

\textbf{Closed surfaces.} Recent years have seen a surge in the use of Neural Implicit Functions (NIFs) for modeling and reconstructing 3D shapes \cite{xie2022neural}. 
These representations are usually implemented with Multi-Layer Perceptrons (MLP) or Convolutional Neural Networks (CNNs), but differ in the learning task \cite{mescheder2019occupancy, chen2019learning, peng2020convolutional, chibane2020implicit}.
On the one hand, some methods \cite{mescheder2019occupancy, chen2019learning, peng2020convolutional, mi2020ssrnet, chibane2020implicit} learn an indicator function or binary occupancies which are used to reconstruct the 3D surface.
On the other hand, alternative methods focus on estimating the SDF at any point in 3D space \cite{park2019deepsdf, jiang2020local, michalkiewicz2019deep, davies2020effectiveness, duan2020curriculum, michalkiewicz2019implicit}.
The landscape of NIFs has further expanded with the introduction of novel approaches like implicit moving least-squares surfaces \cite{liu2021deep}, a differentiable Poisson solver \cite{peng2021shape}, a complex Gabor wavelet \cite{saragadam2023wire}, and a level set alignment loss \cite{ma2023towards}.
Notably, central works in the field \cite{gropp2020implicit,sitzmann2020implicit} approach the problem as finding solutions to the \textit{Eikonal} equation alongside expressive boundary conditions, building upon the literature on solving PDEs with neural networks \cite{sirignano2018dgm, raissi2017physics}. 
In particular, SIREN \cite{sitzmann2020implicit} achieves remarkable results utilizing periodic activation functions, allowing to successfully control the function's differential fields. 
Despite these advancements, a fundamental limitation of current NIF approaches remains in their inability to represent open surfaces, a characteristic often exhibited by real-world objects like scene walls, clothing, or vehicles with inner structures.

\textbf{Open surfaces.} In order to model general non-watertight surfaces, Chibane et al. introduced the idea of learning UDFs, pioneering the handling of open surfaces with neural networks \cite{chibane2020neural}. 
Following this, several methods have aimed to improve the performance of open-surface representation using neural networks \cite{venkatesh2021deep, wang2022rangeudf, ye2022gifs, yang2023neural, zhao2021learning, zhou2022learning}.
For instance, GIFS \cite{ye2022gifs} and Neural Vector Fields \cite{yang2023neural} model the relationships between every two points instead of the relationships between points and surfaces. 
HSDF separately learns sign and distance fields to handle surfaces with arbitrary topologies\cite{wang2022hsdf}.
NeuralUDF \cite{long2023neuraludf} and NeUDF \cite{liu2023neudf} focus on learning an unsigned distance field by volume rendering for multi-view reconstruction of surfaces with arbitrary topologies.
Closer to our work is CAP-UDF \cite{zhou2022learning}, 
that optimizes models on raw point clouds by learning to move 3D point queries until reaching the surface with a field consistency constraint. 
Despite these developments, since UDFs are not differentiable at the zero level set, the gradient field is ill-defined near the isosurface. 
This leads to difficulties during training, but specially during surface reconstruction, where gradients are often used to determine surface presence between points \cite{guillard2022meshudf, zhou2022cap-udf}, compute surface normals, and project points onto the isosurface \cite{chibane2020neural} through gradient descent, among others.
Recent work by Zhou et al. aimed to solve this issue by introducing constraints at the zero level set in the form of losses leading to a smoother surface \cite{zhou2023learning}. 
In contrast, our approach formulates an \textit{Eikonal} problem which is solved by learning a scaled distance function that remains differentiable close to the surface. 
We demonstrate that our method not only enhances reconstruction quality, addressing the prevailing challenges associated with the non-differentiability of the unsigned distance, but also improves \emph{state-of-the-art} in terms of training time performance.

\section{Proposed Approach}
\label{sec:proposed_approach}
\subsection{Mathematical background}

Methods addressing closed surfaces approach the learning of SDFs as finding the solution to a system of Partial Differential Equations (PDEs) governed by the homogeneous \textit{Eikonal} equation with \textit{Dirichlet} and \textit{Neumann} boundary conditions. 
Formally, given a closed surface $\mathcal{S}$ in $\mathcal{C} \subseteq \mathbb{R}^3$ they seek to find a continuous function $f$ that satisfies $\left \| \nabla f \right \| = 1$, with boundary conditions $f_{\rvert_\mathcal{S}} = 0$, $\nabla f_{\rvert_\mathcal{S}} = \mathbf{n}_\mathcal{S}$. Where $\mathbf{n}_\mathcal{S}$ denotes the unitary normal field at the surface. 
Although SDFs are not differentiable at every point and represent only a weak solution to the \textit{Eikonal} equation, recent work has demonstrated considerable success in solving this problem with continuously differentiable implicit neural representation networks \cite{sitzmann2020implicit, gropp2020implicit ,novello2022exploring, clemot2023neural_skeleton}.
These architectures use periodic activations, facilitating smoother optimization processes and improved control over the solution's gradient field.
This is achievable because locations where the signed distance lacks differentiability are distant from the isosurface, hence the approximation tends to be good in a close neighborhood of the zero level set.
In particular, these networks enable accurate computation of critical topological properties, including mean and Gaussian curvature \cite{novello2022exploring}. 
This contrasts with networks featuring piecewise linear activations, such as \textit{ReLUs}, which have null second-order derivatives \cite{sitzmann2020implicit}.

In the context of representing UDFs, these functions cannot be a solution to the \textit{Eikonal} equation both in the interior and exterior regions of the surface, without losing differentiability at the zero level set. 
This presents challenges for continuously differentiable implicit neural representation networks in achieving satisfactory outcomes, being that the approximation errors happen at the isosurface where accuracy is paramount. 
To our knowledge, there has been no successful report of a solution to the \textit{Eikonal} equation accurately approximating unsigned distance functions.

\begin{figure}[t]
  \centering
  \includegraphics[width=1.0\linewidth]{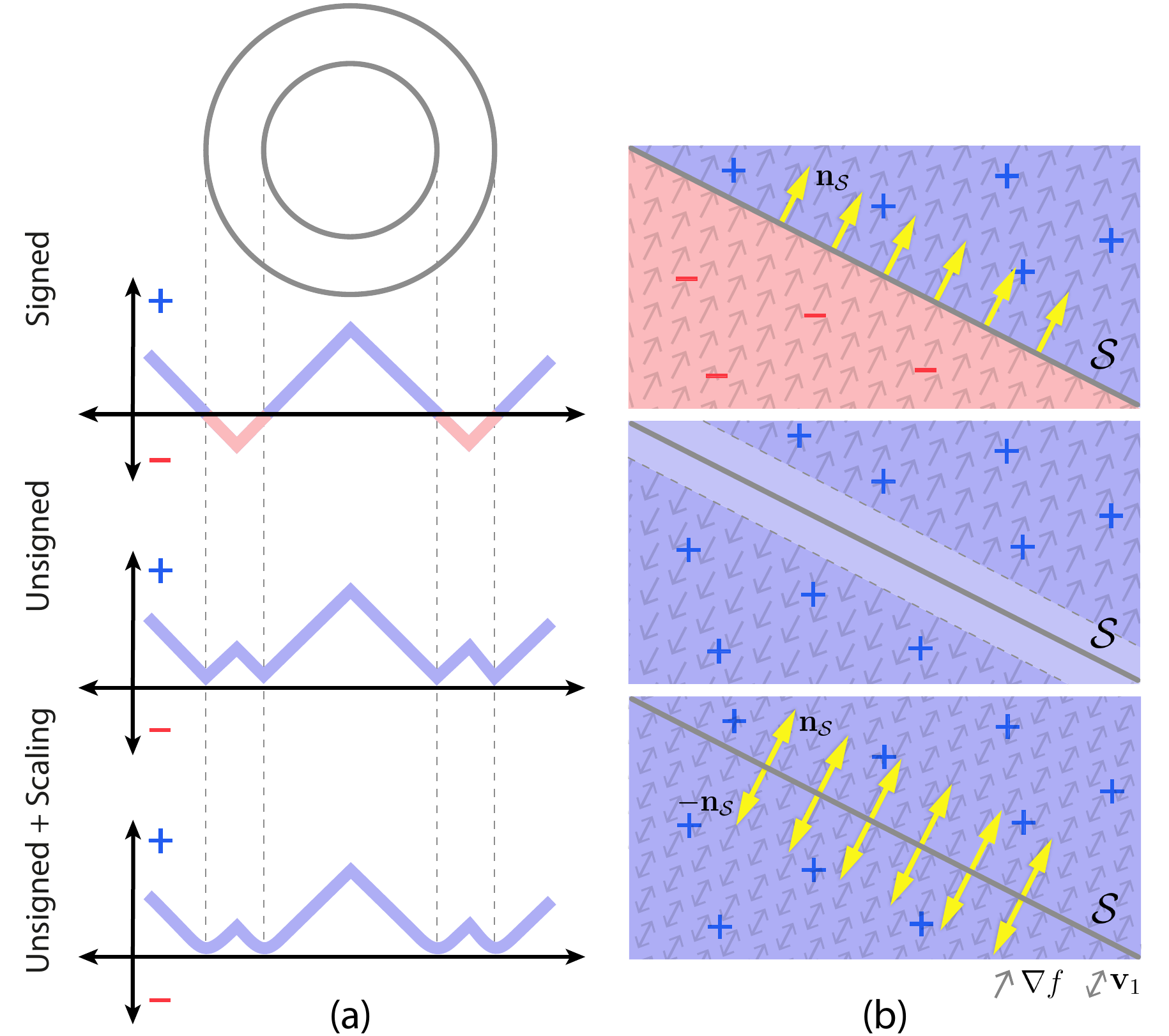}
   \caption{In (a), view of different distance fields for a 2D slice of a torus. Note the effect of hyperbolic scaling near the isosurface (bottom). In (b), sign and gradient for each distance field $f$. In signed distance, the gradient $\nabla f$ at the isosurface is equal to the surface's normal field $\mathbf{n}_\mathcal{S}$ (top). In unsigned distance, the gradient is undefined at the isosurface due to the change in orientation (middle). Our maximum curvature field vectors $\mathbf{v}_1$ align with the surface's unoriented normals (bottom). 
   }
   \label{fig:problem_statement}
\end{figure}

\subsection{Problem statement}
Our first insight is to redefine the unsigned distance field through the application of a hyperbolic scaling. 
In light of this understanding, we propose to learn the parameters $\theta$ of a neural network $f_{\theta}$ with periodic activation functions \cite{sitzmann2020implicit} to approximate the function:

\begin{equation}
    \label{eq:ts}
    \mathit{t_\mathcal{S}}(\mathbf{x})=\mathit{d}_\mathcal{S}(\mathbf{x}) \tanh( \alpha \mathit{d}_\mathcal{S}(\mathbf{x})),
\end{equation}
where $\mathit{d}_\mathcal{S}$ is the unsigned euclidean distance to surface $\mathcal{S}$; and $\alpha$ is a constant value. 
The function $\mathit{t}_\mathcal{S}$ is a differentiable approximation of $\mathit{d}_\mathcal{S}$, whose zero level set is surface $\mathcal{S}$. 
Fig.~\ref{fig:problem_statement}~(a) illustrates hyperbolic scaling's quadratic smoothing near the isosurface and the linear behavior in the distance.
The parameter $\alpha$ controls the distance at which the smoothing occurs.

Our second insight is that the distance scaling enables the application of continuously differentiable implicit neural representation networks to solve an \textit{Eikonal} equation, while retaining the UDF's open surface representation capabilities. 
For this task, we aim to address a heterogeneous \textit{Eikonal} equation, for which we know $t_\mathcal{S}$ (Eq.~\ref{eq:ts}) is a weak solution:

\begin{equation}
    \left\{\begin{matrix}
        \left \| \nabla f \right \| =  \phi  \\ 
        f_{\rvert_\mathcal{S}} = 0 \\
        \nabla f_{\rvert_\mathcal{S}} = \mathbf{0}
    \end{matrix}\right.
    \label{eq:udf_pde}
\end{equation}
where $\phi$ is defined as the L2 norm of the gradient of Eq. \ref{eq:ts}, formally:
{\small
\begin{equation}
    \phi(\mathbf{x})= \tanh(\alpha \mathit{d}_\mathcal{S}(\mathbf{x}))+\alpha \mathit{d}_\mathcal{S}(\mathbf{x})(1-\tanh^2(\alpha \mathit{d}_\mathcal{S}(\mathbf{x}))).
\end{equation}
}

This formulation assumes that the domain boundary is defined by a closed surface $\mathcal{S}$, where \textit{Dirichlet} and \textit{Neumann} boundary conditions can be imposed. 
However, that framework does not directly apply to open surfaces, which do not enclose a well-defined domain and thus cannot support the imposition of boundary constraints. 
To address this limitation, we reformulate the problem as an initial value problem where the \textit{Dirichlet} condition on $\mathcal{S}$ prescribes the initial values for $f$, and where the \textit{Neumann} condition can be omitted given that is already satisfied by the \textit{Eikonal} equation when $\phi(\mathbf{s}) = 0$ for $\mathbf{s} \in \mathcal{S}$. 

The problem outlined in Eq.~\ref{eq:udf_pde} does not incorporate the information of the surface normals.
This contrasts with prior research focused on closed surfaces, where leveraging such information has led to improved accuracy and enhanced reconstruction quality by further constraining the set of feasible solutions.
Our third key insight is that although the gradient of $t_\mathcal{S}$ (Eq. \ref{eq:ts}) becomes null at the isosurface, the direction in which the gradient norm increases most rapidly remains non-null.
This direction is indicated by the maximum curvature of $f$, defined as the eigenvector associated with the largest eigenvalue of the Hessian matrix.
Note that in this context, the maximum curvature does not refer to the surface's curvature, but rather to the curvature of the hyperbolic scaled unsigned distance field $f$, where the function's rate of change varies most rapidly.
Formally, $\mathbf{H}_{t_\mathcal{S}}$ is the Hessian matrix of $t_\mathcal{S}$ which has 3 real eigenvalues $\left | \lambda_1 \right | \geq \left | \lambda_2 \right | \geq \left | \lambda_3 \right |$; let $\mathbf{v_1}$ be a unitary eigenvector associated with $\lambda_1$ at $\mathbf{s}$, it can be shown that $\mathbf{v_1} = \pm \mathbf{n}_\mathcal{S}$ for every $\mathbf{s} \in \mathcal{S}$.
Therefore, we further condition the solution to Eq.~\ref{eq:udf_pde} by adding an extra boundary condition enforcing alignment between the unitary directions associated to the maximum curvature field and the unitary normal field of the surface. 

\subsection{Implicit function learning}
Following the aforementioned definitions, our neural networks are trained to minimize the following loss function:
{\small
\begin{equation}
    \mathcal{L} = \lambda_{e} \mathcal{L}_\mathit{Eikonal} + \lambda_{d} \mathcal{L}_\mathit{Dirichlet} + \lambda_{n} \mathcal{L}_\mathit{Neumann} + \lambda_{g} \mathcal{L}_\mathit{MCurv},
    \label{eq:loss}
\end{equation}}
with $\lambda_i$ constant weights controlling the relevance of each term. 
The term which favors a solution to the \textit{Eikonal} PDE is defined as:
\begin{equation}
    \mathcal{L}_\mathit{Eikonal} = \int_{\mathcal{C}} \left | \left \| \nabla f_\theta(\mathbf{x}) \right \|-  \phi(\mathbf{x}) \right |  \quad d\mathbf{x}.
    \label{eq:loss_eik}
\end{equation}
The \textit{Dirichlet} boundary condition loss term controlling function values at the surface is formally defined as:
\begin{equation}
    \mathcal{L}_\mathit{Dirichlet} = \int_{\mathcal{S}} \left | f_{\theta}(\mathbf{x})  \right | \quad d\mathbf{x}.
    \label{eq:loss_dir}
\end{equation}
Similarly to previous works \cite{novello2022exploring, clemot2023neural_skeleton}, we extend this conditioning to points far from the surface ${\mathcal{S}}$. 
This is achieved by computing an approximation of the function $t_\mathcal{S}$ (Eq. \ref{eq:ts}) based on nearest neighbors.
\textit{Neumann's} boundary condition is expressed in the loss term:
\begin{equation}
    \mathcal{L}_\mathit{Neumann} = \int_{\mathcal{S}} \left \| \nabla f_{\theta}(\mathbf{x})  \right \| \quad d\mathbf{x}.
    \label{eq:loss_neu}
\end{equation}
Finally, we ensure the second-order boundary condition by aligning the directions of maximum curvature with the surface normal through the computation of the following integral:
\begin{equation}
    \mathcal{L}_\mathit{MCurv} = \int_{\mathcal{S}} 1 - \left | \mathbf{v_1}(\mathbf{x}) \cdot \mathbf{n}_\mathcal{S}(\mathbf{x})  \right | \quad d\mathbf{x},
    \label{eq:loss_mpc}
\end{equation}
where $\mathbf{v_1}(\cdot)$ is the unitary eigenvector associated with the largest eigenvalue of $\mathbf{H}_{f_{\theta}}(\mathbf{x})$, the Hessian matrix of $f_{\theta}$ at $\mathbf{x}$.
In practice, we approximate all integrals discretely using a data set comprised of tuples  $\{ (\mathbf{x}_i, \mathit{d}_\mathcal{S}(\mathbf{x}_i), \mathbf{n}_\mathcal{S}(\mathbf{x}_i)) \}_{i}$. 
At every training iteration, the gradient and Hessian of the neural network are computed through automatic differentiation.

\subsection{Isosurface refinement}
In pursuit of solutions that exhibit a consistent, near-zero value at the isosurface —essential for smooth reconstructions— we further fine-tune our networks in a second optimization step. 
Formally, we minimize the following loss function: 
\begin{equation}
    \mathcal{L}_{Refinement} = \lambda_{\mu} \rvert \, \mu ({f_\theta}_{\rvert_\mathcal{S}}) \, \rvert + \lambda_{\sigma} \, \sigma ({f_\theta}_{\rvert_\mathcal{S}}),
\end{equation}
where $\mu(\cdot)$ is the mean value and $\sigma(\cdot)$ the standard deviation. 
By minimizing the mean and variance of the learned function values at the isosurface, we ensure they are as close to zero as possible.
This refinement process effectively reduces the oscillations and deviations that can occur at the critical boundary, thereby enhancing the overall quality of the reconstructed surfaces.

\subsection{Normals and curvature computation}
A significant advantage of our formulation over prior efforts \cite{chibane2020neural, zhou2022cap-udf} is the fact that gradient fields and higher-order derivatives are well-defined in the vicinity of the isosurface, allowing the direct computation of topological properties during inference and reconstruction.
This means that we are able to render unsigned distance fields using standard algorithms such as sphere tracing.
However, directly using the gradients of $f_\theta$ at the isosurface as normals for rendering purposes can be unreliable since their norm is close to 0. 
Hence, for a given surface value threshold $\epsilon$, when a point $\mathbf{s}$ is identified such that $f_\theta(\mathbf{s}) < \epsilon$, we determine the normal direction to the surface by calculating the unit eigenvectors associated with the maximum eigenvalue of $ \mathbf{H}_{f_\theta}(\mathbf{s})$.
Note that there are two unitary eigenvectors per surface point, $\mathbf{v_1}$ and $-\mathbf{v_1}$; however this is usually the desired behavior since open surfaces may not be orientable, such as the \textit{möbius strip}. 
To address this ambiguity in the sign we consider the position of the camera when rendering. 
Additionally, other relevant topological properties such as the mean and Gaussian curvature are also available by means of the maximum unit eigenvector field.
See Sec. \ref{sec:curvature} for more details.

\section{Results and evaluation}
\label{sec:experiments}
\subsection{Experimental setup}
\label{seq:exp_setup}
We conducted a series of experiments to assess the performance of our method. 
To target a broad amount of surfaces, we experimented on three well-known data sets: ShapeNet cars \cite{shapenet2015}, Multi-Garment \cite{bhatnagar2019mgn}, and DeepFashion \cite{zhu2020deepfashion}.
For each data set we trained individual networks on 30 randomly selected oriented point clouds.
For the Deep Fashion and Multi-Garment dataset, we sampled 100,000 surface points, while for ShapeNet cars we doubled the sample size to account for the higher degree of complexity. 
On each training iteration, the loss was computed on a subset of 30,000 points, equally distributed into three distinct groups: surface points (1), far domain points (2), and near domain points (3). 
Surface points (1) were randomly selected from the ground truth point cloud. 
Far points (2) were uniformly generated within the function's domain: a cube with side length 2 and centered at the origin.
All shapes were uniformly scaled to fit within this volume before sampling.
The distance to the point clouds was approximated using a tree-based search algorithm \cite{zhou2018open3d}.
Finally, since we found that a biased near-the-surface sampling improved accuracy, we used near domain points (3) during training.
We constructed this set by randomly displacing the sampled surface points in their normal direction.  
Displacements were sampled from a normal distribution centered at zero with a standard deviation of $0.01$. 
The corresponding unsigned distance was approximated with the distance to the undisplaced surface point. 
This was experimentally found to be sufficiently accurate and fast.

\subsection{Network architecture}

Our network takes spatial coordinates $x,y,z$ as input, and outputs the predicted $t_\mathcal{S}$ value from Eq. \ref{eq:ts}. 
The architecture consists of an 8-layer 256-units MLP with \emph{sine} activations. 
The same number of layers and parameter count were used for baseline methods. 
Loss weights were experimentally set to $ \lambda_{e},\lambda_{d},\lambda_{n}=1e^4$, $\lambda_{g}=1e^3$ and $\lambda_{\mu},\lambda_{\sigma}=1e^5$.
Models were implemented in Python using Pytorch library, and distance metrics were computed using Pytorch3D \cite{ravi2020pytorch3d}.
We trained each network for 3,000 iterations, the first third of them utilized a learning rate of $1\mathrm{e}^{-4}$, in the following third we lowered it to $1\mathrm{e}^{-5}$, and in the final iterations (only the refinement loss active) we set the learning rate to $1\mathrm{e}^{-7}$ with a \textit{cosine} decay. 
These final small learning rates allowed the refinement of the isosurface without losing the gained accuracy in the rest of the field. 
Regarding $\alpha$ (Eq. \ref{eq:ts}), we experimentally found that choosing a value of 100 kept a good balance between training accuracy and reconstruction quality.
Experiments were run on Ubuntu, an Intel(R) Xeon(R) Silver 4310 CPU, 256Gb of RAM, and 2 Nvidia A100 graphics cards of 80Gb VRAM.

\subsection{Mesh reconstruction}

For a thorough comparison of our method against existing techniques, we undertake mesh reconstruction using two gradient-based Marching Cubes algorithms, referred here as MC1 \cite{zhou2022cap-udf} and MC2 \cite{guillard2022meshudf}.
These methods perform linear interpolation between grid values for accurate placement of triangle vertices. 
Given this requirement and considering that the function $t_\mathcal{S}$, as described in Equation \ref{eq:ts}, exhibits quadratic behavior near the isosurface, we retrieve the true unsigned distance by first dividing by $\alpha$ and then applying the square root. We present results using the Chamfer distance (L1,L2), pervasive in related literature. Additionally, we compute the Normal Consistency metric (NC) as the mean value of the absolute \textit{cosine} similarities between each point to its closest neighbor in the other point cloud.

\subsection{Surface reconstruction}
\subsubsection{Closed shapes}

\begin{table}[b!]
    \centering
    \resizebox{\columnwidth}{!}{%
    \begin{tabular} {  c  c c c  } 
    \toprule
    Method & time(s) $\downarrow$ & L1CD $\downarrow$ & L2CD $\downarrow$ \\
    \hline
    DeepSDF \cite{park2019deepsdf} & $\mathbf{94}$ & $19.40$ & $0.206$ \\ 
    SIREN \cite{sitzmann2020implicit} & $379$ & $15.40$ & $0.171$ \\ 
    CAP-UDF \cite{zhou2022cap-udf} & $1080$ & $\mathbf{9.48}$ & $0.030$ \\ 
    \hline
    Ours (MC1 \cite{zhou2022cap-udf}) & \multirow{2}{1em}{$319$} & $\mathbf{9.48}$ & $\mathbf{0.028}$ \\
    Ours (MC2 \cite{guillard2022meshudf}) & & $9.49$ & $\mathbf{0.028}$ \\
    \bottomrule
    \end{tabular}}
    \caption{Training time, L1, and L2 mean Chamfer distances ($\times10^3$) for the closed ShapeNet cars data set. DeepSDF and SIREN were trained on signed distances, while CAP-UDF and our method were trained on unsigned distances.}
    \label{tab:closed}
\end{table}

We benchmarked DUDF against state-of-the-art methods for representing closed surfaces, where signed distances are well-defined. 
Given that our primary focus is not on closed surface representation, we limited this comparison to the ShapeNet car dataset \cite{shapenet2015}, modifying the meshes for closure and omitting internal structures \cite{xu2019disn}. 
The comparative results are detailed in Table \ref{tab:closed}. 

While DeepSDF \cite{park2019deepsdf} excels in speed, it falls short in reconstruction quality. Being primarily designed for shape generation, DeepSDF is outperformed by methods tailored for accurate single shape representation.
\begin{figure*}[ht]
  \centering
  \includegraphics[width=1.0\linewidth]{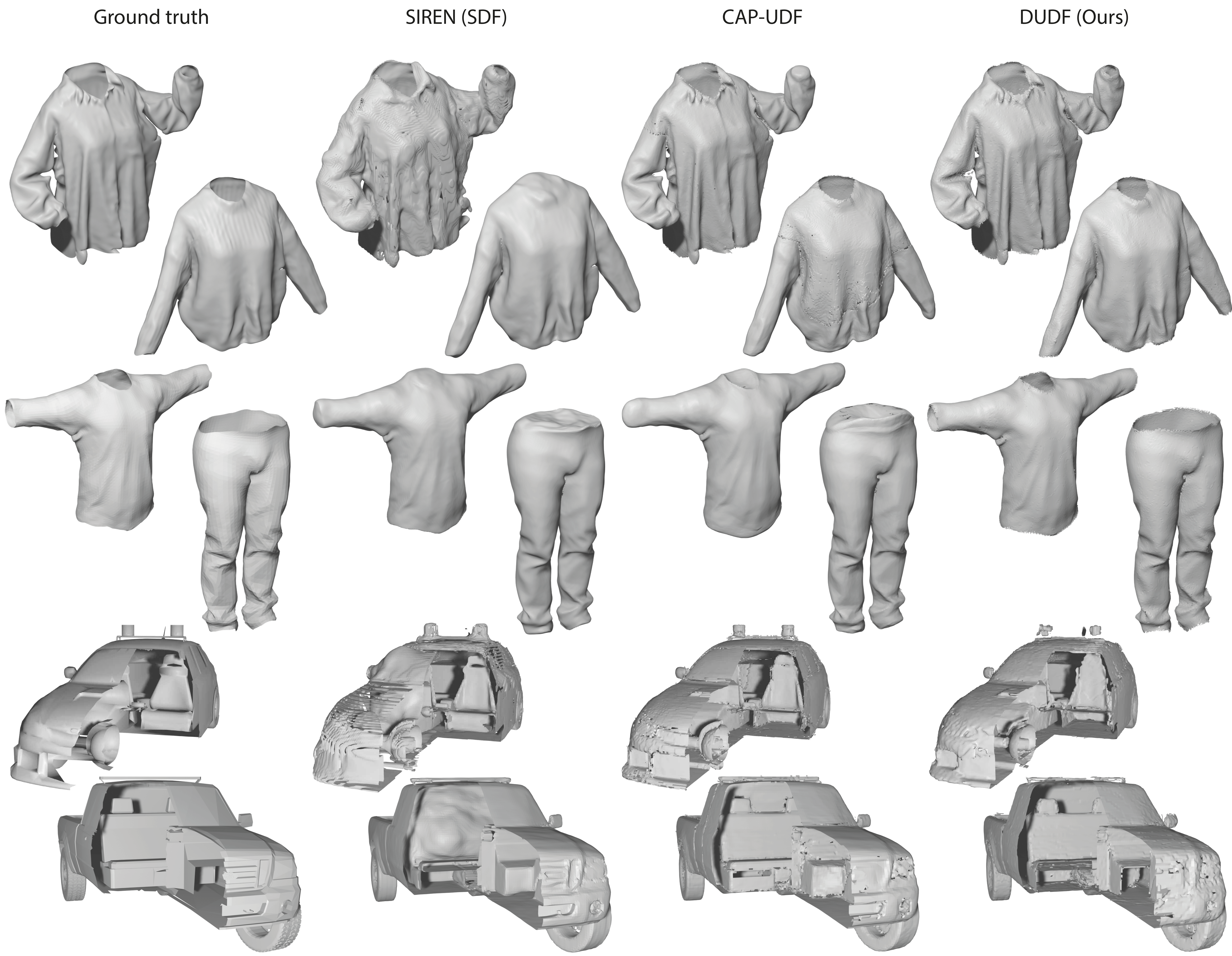}
   \caption{Comparisons on DeepFashion \cite{zhu2020deepfashion} (top row), Multi-Garment \cite{bhatnagar2019mgn} (middle row), and ShapeNet cars \cite{shapenet2015} (bottom row) data sets. DUDF preserves fine details and accurately represents complex geometries without closing holes, outperforming SIREN (SDF), which tends to smooth and round models, and CAP-UDF, which captures sharp features but often closes open surfaces. Reconstructions for CAP-UDF and DUDF performed with MC1 \cite{zhou2022cap-udf}.}
   \label{fig:comparison}
\end{figure*}
Despite SIREN \cite{sitzmann2020implicit} not reporting performance on learning distance fields, we included it in our comparison for the sake of fairness and comprehensiveness, given
that our method’s backbone is built upon its architecture.
When tested on close surfaces, SIREN shows improved shape representation with enough detail and smoothness.
However, since its zero level set satisfies the implicit function theorem, it cannot properly represent shapes with right angles and thin structures.
In contrast, CAP-UDF \cite{zhou2022cap-udf} can better approximate these sharp features because it does not rely on continuously differentiable NNs, thereby achieving higher accuracy. 
However, the training time is high and the reliance on gradient based Marching Cubes methods for surface reconstruction tends to create surfaces with small holes and discontinuities which hinders the visual quality in some cases.
Our approach captures a differentiable function's zero level set like SIREN, but uniquely bypasses the implicit function theorem by having null gradients at the isosurface. 
This allows our method to effectively represent sharp features and complex geometries, leading to improved precision in surface reconstruction, particularly in scenarios where surfaces are not inherently smooth.

\subsubsection{Open surfaces}
\label{sec:open-surf}

When evaluating open-surface representation, we primarily compare our method against CAP-UDF \cite{zhou2022cap-udf}. 
This approach was shown to be more effective than earlier techniques (i.e. NDF \cite{chibane2020neural}), that required a complex process for creating a dense point cloud and reconstructing a triangulated surface using the Ball Pivoting algorithm \cite{bernandini1999ball}. 
This often leads to poor quality meshes with holes and irregularities.
Additionally, a recent method by Zhou et al. \cite{zhou2023learning} has introduced constraints on the zero level set, which the authors claim it leads to better reconstructions. 
However, the absence of available implementation code in their official repository precluded a fair comparison with our method.
Therefore, we concentrate our analysis on the most recent and available methodologies.

\begin{table*}[!t]
    \centering
    \resizebox{\textwidth}{!}{%
    \begin{tabular} {  c c c c } 
    \toprule
        \begin{tabular}{c}
            \multirow{2}{3em}{Method}  \\ \\
            \hline
            SIREN \cite{sitzmann2020implicit} \\
            CAP-UDF \cite{zhou2022cap-udf} \\
            \hline
            Ours (MC1 \cite{zhou2022cap-udf}) \\
            Ours (MC2 \cite{guillard2022meshudf})
        \end{tabular} &

        \begin{tabular}{c c c c}
            \multicolumn{4}{c}{DeepFashion \cite{zhu2020deepfashion}} \\
            time(s) $\downarrow$ & L1CD $\downarrow$ & L2CD $\downarrow$ & NC $\downarrow$\\
            \hline
            $376$ & $27.3$ & $1.980$ & $0.107$ \\
            $1390$ & $18.1$ & $1.110$ & $0.080$ \\
            \hline
            \multirow{2}{1.5em}{$\mathbf{326}$} & $\mathbf{9.01}$ & $\mathbf{0.025}$ & $0.024$ \\
             & $9.14$ & $0.027$ & $\mathbf{0.020}$\\
        \end{tabular}&

        \begin{tabular}{c c c c}
            \multicolumn{4}{c}{Multi-Garment \cite{bhatnagar2019mgn}} \\
            time(s) $\downarrow$ & L1CD $\downarrow$ & L2CD $\downarrow$ & NC $\downarrow$\\
            \hline
            $374$ & $40.5$ & $8.810$ & $0.094$\\
            $1440$ & $18.5$ & $1.190$ & $0.083$ \\
            \hline
            \multirow{2}{1.5em}{$\mathbf{318}$} & $\mathbf{8.70}$ & $\mathbf{0.024}$ & $0.026$\\
             & $8.82$ & $0.026$ & $\mathbf{0.021}$\\
        \end{tabular} &

        \begin{tabular}{c c c c}
            \multicolumn{4}{c}{ShapeNet cars \cite{shapenet2015}} \\
            time(s) $\downarrow$ & L1CD $\downarrow$ & L2CD $\downarrow$ & NC $\downarrow$\\
            \hline
            $751$ & $16.9$ & $0.170$ & $\mathbf{0.240}$\\
            $1040$ & $\mathbf{10.9}$ & $0.071$ & $0.288$ \\
            \hline
            \multirow{2}{1.5em}{$\mathbf{317}$} & $12.3$ & $\mathbf{0.057}$ & $0.387$ \\
             & $13.7$ & $0.081$ & $0.304$ \\
        \end{tabular} \\
        
        \bottomrule
    \end{tabular}
    }
    \caption{Training time, L1 and L2 mean Chamfer distances ($\times10^3$), and Normal Consistency (NC) for the evaluated open surface data sets. SIREN was trained as described in the original paper, while CAP-UDF and our method were trained on unsigned distances.}
    \label{tab:open}
\end{table*}

Qualitative comparisons can be observed in Fig. \ref{fig:comparison} and a full quantitative analysis is presented in Table \ref{tab:open}. 
On the one hand, our method demonstrates a significant improvement in efficiency, consuming up to an order of magnitude less computational time than CAP-UDF, while also showing enhanced performance across all three data sets. 
This advantage is particularly evident in the DeepFashion and Multi-Garment data sets, where CAP-UDF tends to inaccurately close openings (like those at the ends of sleeves in clothing). 
On the other hand, SIREN creates a negative distance valued shell around the isosurface.
This causes to either close every hole or a thickened surface enclosing the ground-truth level set. 

\begin{figure}[b]
  \centering
  \includegraphics[width=1.0\linewidth]{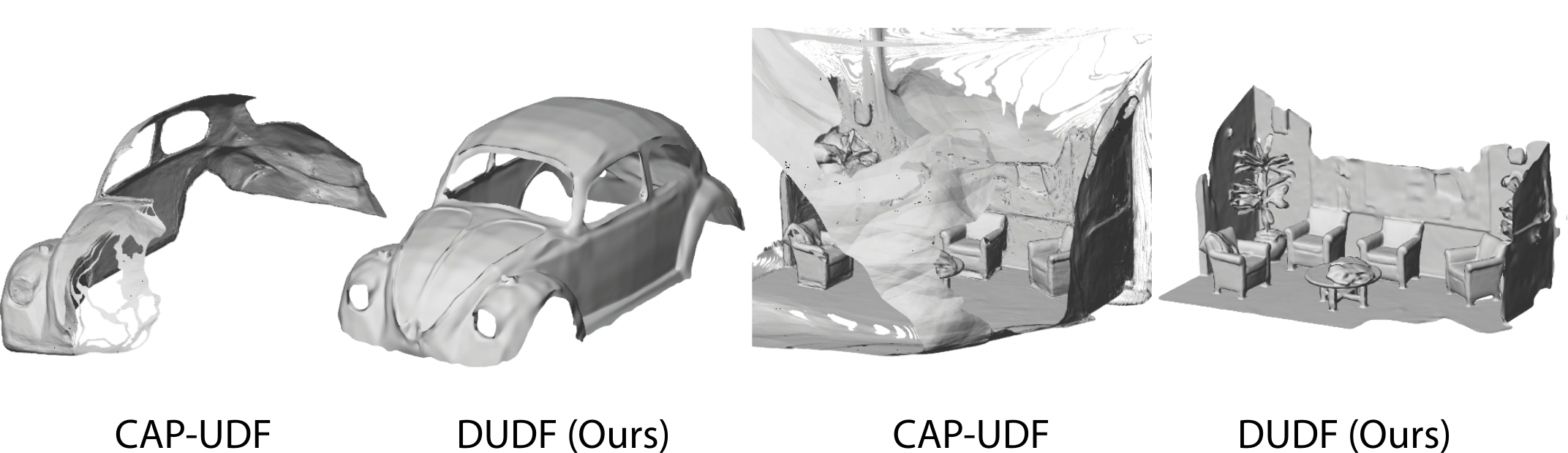}
   \caption{Rendering examples using sphere tracing. CAP-UDF struggles with non-linear growth of their unsigned distance fields, causing the sphere tracing marching step to miss the surface. Conversely, our method demonstrates precision in direct rendering scenarios.}
   \label{fig:sphere_tracing}
\end{figure}

In addition to these experiments, we compare our method with CAP-UDF in the context of sphere tracing rendering.
This comparison is presented in Fig. \ref{fig:sphere_tracing}, where we showcase two rendering examples. 
Since learned unsigned distance fields in CAP-UDF do not grow linearly away from the surface, the marching steps in sphere tracing often fail to accurately intersect the surface, leading to undesirable visual artifacts. 
Even when hitting the surface, gradients might be undefined, leading to noisy images.
In contrast, our proposed distance function does not exhibit these problems, offering a more robust framework for direct rendering. 
Furthermore, our ability to compute normals using the maximum curvature field is useful for shading purposes, enhancing the rendering quality without undergoing an intermediate 3D mesh reconstruction step.

\subsection{Ablation study}

We now dissect the impact of individual loss components on the network's accuracy through an ablation study. 
The methodology involves selectively deactivating terms in our loss function (Eq. \ref{eq:loss}) to isolate their contributions. 
We adopted the experimental framework from the prior section, training on 30 meshes from the DeepFashion dataset \cite{zhu2020deepfashion}. 
We report metrics for reconstructions performed using MC2 \cite{guillard2022meshudf}, yielding results analogous to those obtained with MC1 \cite{zhou2022cap-udf}.

The findings presented in Table \ref{tab:ablation} underscore the significance of the \textit{Eikonal} equation (Eq. \ref{eq:loss_eik}) in achieving superior normal consistency ($\lambda_g$).
Moreover, the refinement step during training ($\lambda_\mu, \lambda_\sigma$) is an important element for normal consistency improvement, offering substantial increased performance for a small additional computational cost.
Although $\mathcal{L}_{MCurv}$ (Eq. \ref{eq:loss_mpc}) does not offer substantial quantitative benefits ($\lambda_g$), we found that this loss term becomes very relevant qualitatively when rendering through direct methods (see Supp. Material for more details).  Additionally, we contrasted the effects of approximating the ground truth distance function $d_\mathcal{S}$ (with its distinct \textit{Eikonal} problem), using the same network architecture and sampling scheme, with \textit{sine} and \textit{ReLU} activations.
The findings from this study show that such an approach leads to suboptimal outcomes, thereby underscoring the effectiveness of $t_\mathcal{S}$ and the derived \textit{Eikonal} problem proposed.

Finally, we ablated parameter $\alpha$ (see Supp. Material). 
As $\alpha$ gets larger, function $t_\mathcal{S}$ closely approximates $d_\mathcal{S}$, increasing the reconstruction error (probably due to the non-differentiability at the isosurface). 
Smaller $\alpha$ values enlarge the quadratic strip near the isosurface, which is harder to supervise effectively and hinders the performance of MC.

\begin{table}[hb!]
    \centering
    \begin{tabular} {  c c c c c } 
    \toprule
    Method & time(s) $\downarrow$ & L1CD $\downarrow$ & L2CD $\downarrow$ & NC $\downarrow$ \\
    \hline
    Baseline & $326$ & $\mathbf{9.14}$ & $\mathbf{0.027}$ & $\mathbf{0.020}$ \\
    $\lambda_e = 0$ & $312$ & $9.43$ & $0.028$ & $0.033$ \\
    $\lambda_g = 0$ & $150$ & $9.16$ & $\mathbf{0.027}$ & $0.021$ \\
    $\lambda_e,\lambda_g = 0$ & $\mathbf{109}$ & $9.44$ & $\mathbf{0.027}$ & $0.035$ \\
    $\lambda_\mu, \lambda_\sigma = 0$ & $302$ & $9.24$ & $\mathbf{0.027}$ & $0.031$ \\
    $d_\mathcal{S}$ (sine) & $150$ & $31.2$ & $0.830$ & $0.057$ \\
    $d_\mathcal{S}$ ($ReLU$) & $145$ & $47.5$ & $2.500$ & $0.149$ \\
    \bottomrule
    \end{tabular}
    \caption{Quantitative impact of each loss component on the reconstruction accuracy of our network. We report training time, L1, and L2 mean Chamfer distances $\times10^3$, and Normal Consistency (NC). Last two rows correspond to approximating the ground truth distance function $d_\mathcal{S}$ instead of $t_\mathcal{S}$.}
    \label{tab:ablation}
\end{table}

\subsection{Computing mean and Gaussian curvature} \label{sec:curvature}
A significant advantage of our method over previous approaches is the possibility to compute curvatures. 
Previous work learned non-differentiable functions, which preclude the direct computation of geometrical properties such as mean and Gaussian curvature. 
Our differentiable UDFs facilitates the extraction of these curvature values. 
Leveraging the divergence ($\nabla \cdot$) of the normal field ($\mathbf{n}_\mathcal{S}$), the mean curvature $H$ at any point on the surface $\mathbf{s}$ is then computed \cite{goldman2005curvature, do2016differential, novello2022exploring} as: 
\begin{equation}
    H(\mathbf{s}) = \frac{1}{2} \nabla \cdot \mathbf{n}_\mathcal{S}(\mathbf{s}).
\end{equation}

\begin{figure}[b]
  \centering
  \includegraphics[width=1.0\linewidth]{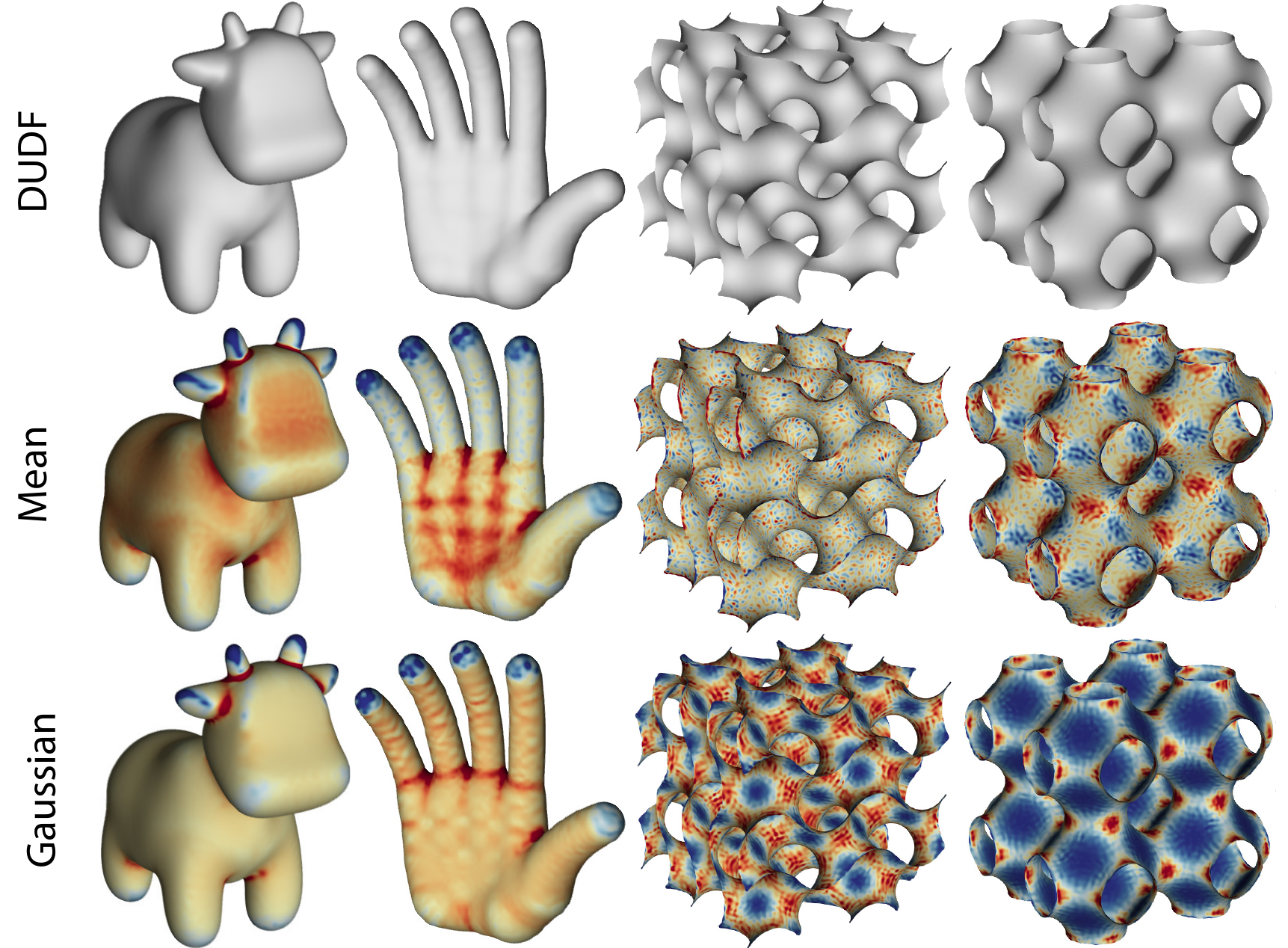}
   \caption{Mean and Gaussian curvatures computed with our method for closed surfaces (left) and open surfaces (right).
   }
   \label{fig:curvatures}
\end{figure}

Additionally, the Gaussian curvature $K$ can be computed using the determinant of a matrix composed of the Jacobian matrix $J_{\mathbf{n}_\mathcal{S}}(\mathbf{s})$ of the unit normal field ($\mathbf{n}_\mathcal{S}$) along with the normal field itself. 
Formally, we compute:
\begin{equation}
    K(\mathbf{s}) = -\det\left[ 
        \begin{array}{c|c} 
          J_{\mathbf{n}_\mathcal{S}}(\mathbf{s}) & \mathbf{n}_\mathcal{S}(\mathbf{s}) \\ 
          \hline 
          \mathbf{n}_\mathcal{S}(\mathbf{s})^t & 0 
        \end{array} 
        \right].
\end{equation}
In Fig. \ref{fig:curvatures} we show mean and Gaussian curvatures for open and closed surfaces computed with our method.

\section{Limitations and future work}
\label{sec:limitations}
Our method encounters certain limitations. 
Primarily, by approximating the function $t_\mathcal{S}$ (Eq. \ref{eq:ts}), the \textit{Eikonal} equation requires supervision with ground truth unsigned distances, in contrast to signed distance approaches where it is constant everywhere. 
In addition, the quadratic behavior of the function near the isosurface produces a wider near-zero strip which can lead to slightly inflated representations. 
This effect can be lessened by computing square roots, at the expense of an increased computational cost. 
Besides hyperbolic scaling, we experimented with the square distance and \emph{SmoothL1}. 
The former requires direct rendering to compute square roots at every marching step and \emph{SmoothL1} does not yield the true distance for points far from the isosurface when $\delta$ is nonzero. 
This is not the case for our hyperbolic function, that satisfies $t_\mathcal{S} = d_\mathcal{S}$ for distant domain points. Future research can find other suited functions for distance field learning.
Secondly, we observed that surfaces extracted using gradient-based Marching Cubes methods can sometimes be less smooth than directly rendering the function using sphere tracing. 
This is likely due to the current reconstruction algorithm's sensitivity to noise, an issue that might be addressed by employing a tangency-aware surface reconstruction strategy \cite{sellan2023reach}.
Additionally, computing the maximum curvature direction field is computationally expensive, requiring three network passes and the diagonalization of the Hessian matrix. 
Future network's designs could reduce the number of parameters, thus lessening the computational burden associated with computing topological features.

\section{Conclusion}
\label{sec:conclusion}
In this work we introduced Differentiable Unsigned Distance Fields (DUDF) with Hyperbolic Scaling, a novel approach that addresses inherent limitations of traditional UDFs in representing open surfaces. 
By applying a hyperbolic transformation to the distance field, we define a new variant of the \textit{Eikonal} problem, tailored with unique boundary conditions.
This adaptation enables our model to capitalize on the robust framework of continuously differentiable neural networks, thus enhancing reconstruction quality and training efficiency.
The conducted experiments provide evidence that our approach can lead to improvements in reconstruction quality and computational performance when compared to several state-of-the-art methods. 
Moreover, the ability to accurately calculate topological features such as normals and curvatures is a notable benefit of our model, offering additional utility in geometric processing tasks and rendering.
While our results are promising, showing better accuracy and efficiency in most cases, we acknowledge the need for further research and development. 
The potential applications of DUDF in 3D geometry processing are broad, and we are optimistic about its utility in practical scenarios. 

\section{Aknowledgments}
This project was supported by grants from Salesforce, USA (EinsteinAI 2020), National Scientific and Technical Research Council (CONICET), (PIP 2021-2023 GI -11220200102981CO), and Universidad Torcuato Di Tella, Argentina. We also thank Facundo Fainstein and Tomas Delgado for valuable discussions on the \textit{Eikonal} problem.

{\small
\bibliographystyle{ieee_fullname}
\bibliography{egbib}

\begin{thebibliography}{10}\itemsep=-1pt

\bibitem{bernandini1999ball}
Fausto Bernardini, J. Mittleman, Holly Rushmeier, Cláudio Silva, and Gabriel
  Taubin.
\newblock The ball-pivoting algorithm for surface reconstruction.
\newblock {\em Visualization and Computer Graphics, IEEE Transactions on},
  5:349 -- 359, 11 1999.

\bibitem{bhatnagar2019mgn}
Bharat~Lal Bhatnagar, Garvita Tiwari, Christian Theobalt, and Gerard Pons-Moll.
\newblock Multi-garment net: Learning to dress 3d people from images.
\newblock In {\em {IEEE} International Conference on Computer Vision ({ICCV})}.
  {IEEE}, oct 2019.

\bibitem{shapenet2015}
Angel~X Chang, Thomas Funkhouser, Leonidas Guibas, Pat Hanrahan, Qixing Huang,
  Zimo Li, Silvio Savarese, Manolis Savva, Shuran Song, Hao Su, et~al.
\newblock Shapenet: An information-rich 3d model repository.
\newblock {\em arXiv preprint arXiv:1512.03012}, 2015.

\bibitem{chen2019learning}
Zhiqin Chen and Hao Zhang.
\newblock Learning implicit fields for generative shape modeling.
\newblock In {\em Proceedings of the IEEE/CVF Conference on Computer Vision and
  Pattern Recognition}, pages 5939--5948, 2019.

\bibitem{chibane2020implicit}
Julian Chibane, Thiemo Alldieck, and Gerard Pons-Moll.
\newblock Implicit functions in feature space for 3d shape reconstruction and
  completion.
\newblock In {\em Proceedings of the IEEE/CVF conference on computer vision and
  pattern recognition}, pages 6970--6981, 2020.

\bibitem{chibane2020neural}
Julian Chibane, Gerard Pons-Moll, et~al.
\newblock Neural unsigned distance fields for implicit function learning.
\newblock {\em Advances in Neural Information Processing Systems},
  33:21638--21652, 2020.

\bibitem{clemot2023neural_skeleton}
Mattéo Clémot and Julie Digne.
\newblock Neural skeleton: Implicit neural representation away from the
  surface.
\newblock {\em Computers \& Graphics}, pages 368--378, 2023.

\bibitem{curless1996volumetric}
Brian Curless and Marc Levoy.
\newblock A volumetric method for building complex models from range images.
\newblock In {\em Proceedings of the 23rd annual conference on Computer
  graphics and interactive techniques}, pages 303--312, 1996.

\bibitem{davies2020effectiveness}
Thomas Davies, Derek Nowrouzezahrai, and Alec Jacobson.
\newblock On the effectiveness of weight-encoded neural implicit 3d shapes.
\newblock {\em arXiv preprint arXiv:2009.09808}, 2020.

\bibitem{do2016differential}
Manfredo~P Do~Carmo.
\newblock {\em Differential geometry of curves and surfaces: revised and
  updated second edition}.
\newblock Courier Dover Publications, 2016.

\bibitem{duan2020curriculum}
Yueqi Duan, Haidong Zhu, He Wang, Li Yi, Ram Nevatia, and Leonidas~J Guibas.
\newblock Curriculum deepsdf.
\newblock In {\em Computer Vision--ECCV 2020: 16th European Conference,
  Glasgow, UK, August 23--28, 2020, Proceedings, Part VIII 16}, pages 51--67.
  Springer, 2020.

\bibitem{goldman2005curvature}
Ron Goldman.
\newblock Curvature formulas for implicit curves and surfaces.
\newblock {\em Computer Aided Geometric Design}, 22(7):632--658, 2005.

\bibitem{gropp2020implicit}
Amos Gropp, Lior Yariv, Niv Haim, Matan Atzmon, and Yaron Lipman.
\newblock Implicit geometric regularization for learning shapes.
\newblock {\em arXiv preprint arXiv:2002.10099}, 2020.

\bibitem{guillard2022meshudf}
Benoit Guillard, Federico Stella, and Pascal Fua.
\newblock Meshudf: Fast and differentiable meshing of unsigned distance field
  networks, 2022.

\bibitem{jiang2020local}
Chiyu Jiang, Avneesh Sud, Ameesh Makadia, Jingwei Huang, Matthias Nie{\ss}ner,
  Thomas Funkhouser, et~al.
\newblock Local implicit grid representations for 3d scenes.
\newblock In {\em Proceedings of the IEEE/CVF Conference on Computer Vision and
  Pattern Recognition}, pages 6001--6010, 2020.

\bibitem{liu2021deep}
Shi-Lin Liu, Hao-Xiang Guo, Hao Pan, Peng-Shuai Wang, Xin Tong, and Yang Liu.
\newblock Deep implicit moving least-squares functions for 3d reconstruction.
\newblock In {\em Proceedings of the IEEE/CVF Conference on Computer Vision and
  Pattern Recognition}, pages 1788--1797, 2021.

\bibitem{liu2023neudf}
Yu-Tao Liu, Li Wang, Jie Yang, Weikai Chen, Xiaoxu Meng, Bo Yang, and Lin Gao.
\newblock Neudf: Leaning neural unsigned distance fields with volume rendering.
\newblock In {\em Proceedings of the IEEE/CVF Conference on Computer Vision and
  Pattern Recognition}, pages 237--247, 2023.

\bibitem{long2023neuraludf}
Xiaoxiao Long, Cheng Lin, Lingjie Liu, Yuan Liu, Peng Wang, Christian Theobalt,
  Taku Komura, and Wenping Wang.
\newblock Neuraludf: Learning unsigned distance fields for multi-view
  reconstruction of surfaces with arbitrary topologies.
\newblock In {\em Proceedings of the IEEE/CVF Conference on Computer Vision and
  Pattern Recognition}, pages 20834--20843, 2023.

\bibitem{lorensen1998marching}
William~E Lorensen and Harvey~E Cline.
\newblock Marching cubes: A high resolution 3d surface construction algorithm.
\newblock In {\em Seminal graphics: pioneering efforts that shaped the field},
  volume~21, pages 347--353. 1998.

\bibitem{ma2023towards}
Baorui Ma, Junsheng Zhou, Yu-Shen Liu, and Zhizhong Han.
\newblock Towards better gradient consistency for neural signed distance
  functions via level set alignment.
\newblock In {\em Proceedings of the IEEE/CVF Conference on Computer Vision and
  Pattern Recognition}, pages 17724--17734, 2023.

\bibitem{mescheder2019occupancy}
Lars Mescheder, Michael Oechsle, Michael Niemeyer, Sebastian Nowozin, and
  Andreas Geiger.
\newblock Occupancy networks: Learning 3d reconstruction in function space.
\newblock In {\em Proceedings of the IEEE/CVF conference on computer vision and
  pattern recognition}, pages 4460--4470, 2019.

\bibitem{mi2020ssrnet}
Zhenxing Mi, Yiming Luo, and Wenbing Tao.
\newblock Ssrnet: Scalable 3d surface reconstruction network.
\newblock In {\em Proceedings of the IEEE/CVF Conference on Computer Vision and
  Pattern Recognition}, pages 970--979, 2020.

\bibitem{michalkiewicz2019deep}
Mateusz Michalkiewicz, Jhony~K Pontes, Dominic Jack, Mahsa Baktashmotlagh, and
  Anders Eriksson.
\newblock Deep level sets: Implicit surface representations for 3d shape
  inference.
\newblock {\em arXiv preprint arXiv:1901.06802}, 2019.

\bibitem{michalkiewicz2019implicit}
Mateusz Michalkiewicz, Jhony~K Pontes, Dominic Jack, Mahsa Baktashmotlagh, and
  Anders Eriksson.
\newblock Implicit surface representations as layers in neural networks.
\newblock In {\em Proceedings of the IEEE/CVF International Conference on
  Computer Vision}, pages 4743--4752, 2019.

\bibitem{newcombe2011kinectfusion}
Richard~A Newcombe, Shahram Izadi, Otmar Hilliges, David Molyneaux, David Kim,
  Andrew~J Davison, Pushmeet Kohi, Jamie Shotton, Steve Hodges, and Andrew
  Fitzgibbon.
\newblock Kinectfusion: Real-time dense surface mapping and tracking.
\newblock In {\em 2011 10th IEEE international symposium on mixed and augmented
  reality}, pages 127--136. Ieee, 2011.

\bibitem{novello2022exploring}
Tiago Novello, Guilherme Schardong, Luiz Schirmer, Vinicius da Silva, Helio
  Lopes, and Luiz Velho.
\newblock Exploring differential geometry in neural implicits, 2022.

\bibitem{park2019deepsdf}
Jeong~Joon Park, Peter Florence, Julian Straub, Richard Newcombe, and Steven
  Lovegrove.
\newblock Deepsdf: Learning continuous signed distance functions for shape
  representation.
\newblock In {\em Proceedings of the IEEE/CVF conference on computer vision and
  pattern recognition}, pages 165--174, 2019.

\bibitem{peng2021shape}
Songyou Peng, Chiyu Jiang, Yiyi Liao, Michael Niemeyer, Marc Pollefeys, and
  Andreas Geiger.
\newblock Shape as points: A differentiable poisson solver.
\newblock {\em Advances in Neural Information Processing Systems},
  34:13032--13044, 2021.

\bibitem{peng2020convolutional}
Songyou Peng, Michael Niemeyer, Lars Mescheder, Marc Pollefeys, and Andreas
  Geiger.
\newblock Convolutional occupancy networks.
\newblock In {\em Computer Vision--ECCV 2020: 16th European Conference,
  Glasgow, UK, August 23--28, 2020, Proceedings, Part III 16}, pages 523--540.
  Springer, 2020.

\bibitem{raissi2017physics}
Maziar Raissi, Paris Perdikaris, and George~Em Karniadakis.
\newblock Physics informed deep learning (part ii): Data-driven discovery of
  nonlinear partial differential equations, 2017.

\bibitem{ravi2020pytorch3d}
Nikhila Ravi, Jeremy Reizenstein, David Novotny, Taylor Gordon, Wan-Yen Lo,
  Justin Johnson, and Georgia Gkioxari.
\newblock Accelerating 3d deep learning with pytorch3d.
\newblock {\em arXiv:2007.08501}, 2020.

\bibitem{saragadam2023wire}
Vishwanath Saragadam, Daniel LeJeune, Jasper Tan, Guha Balakrishnan, Ashok
  Veeraraghavan, and Richard~G Baraniuk.
\newblock Wire: Wavelet implicit neural representations.
\newblock In {\em Proceedings of the IEEE/CVF Conference on Computer Vision and
  Pattern Recognition}, pages 18507--18516, 2023.

\bibitem{sellan2023reach}
Silvia Sell{\'a}n, Christopher Batty, and Oded Stein.
\newblock Reach for the spheres: Tangency-aware surface reconstruction of sdfs.
\newblock {\em arXiv preprint arXiv:2308.09813}, 2023.

\bibitem{sirignano2018dgm}
Justin Sirignano and Konstantinos Spiliopoulos.
\newblock {DGM}: A deep learning algorithm for solving partial differential
  equations.
\newblock {\em Journal of Computational Physics}, pages 1339--1364, 2018.

\bibitem{sitzmann2020implicit}
Vincent Sitzmann, Julien Martel, Alexander Bergman, David Lindell, and Gordon
  Wetzstein.
\newblock Implicit neural representations with periodic activation functions.
\newblock {\em Advances in neural information processing systems},
  33:7462--7473, 2020.

\bibitem{venkatesh2021deep}
Rahul Venkatesh, Tejan Karmali, Sarthak Sharma, Aurobrata Ghosh, R~Venkatesh
  Babu, L{\'a}szl{\'o}~A Jeni, and Maneesh Singh.
\newblock Deep implicit surface point prediction networks.
\newblock In {\em Proceedings of the IEEE/CVF International Conference on
  Computer Vision}, pages 12653--12662, 2021.

\bibitem{wang2022rangeudf}
Bing Wang, Zhengdi Yu, Bo Yang, Jie Qin, Toby Breckon, Ling Shao, Niki Trigoni,
  and Andrew Markham.
\newblock Rangeudf: Semantic surface reconstruction from 3d point clouds.
\newblock {\em arXiv preprint arXiv:2204.09138}, 2022.

\bibitem{wang2022hsdf}
Li Wang, Weikai Chen, Xiaoxu Meng, Bo Yang, Jintao Li, Lin Gao, et~al.
\newblock Hsdf: Hybrid sign and distance field for modeling surfaces with
  arbitrary topologies.
\newblock {\em Advances in Neural Information Processing Systems},
  35:32172--32185, 2022.

\bibitem{xie2022neural}
Yiheng Xie, Towaki Takikawa, Shunsuke Saito, Or Litany, Shiqin Yan, Numair
  Khan, Federico Tombari, James Tompkin, Vincent Sitzmann, and Srinath Sridhar.
\newblock Neural fields in visual computing and beyond.
\newblock In {\em Computer Graphics Forum}, volume~41, pages 641--676. Wiley
  Online Library, 2022.

\bibitem{xu2019disn}
Qiangeng Xu, Weiyue Wang, Duygu Ceylan, Radomir Mech, and Ulrich Neumann.
\newblock Disn: Deep implicit surface network for high-quality single-view 3d
  reconstruction.
\newblock {\em CoRR}, abs/1905.10711, 2019.

\bibitem{yang2023neural}
Xianghui Yang, Guosheng Lin, Zhenghao Chen, and Luping Zhou.
\newblock Neural vector fields: Implicit representation by explicit learning.
\newblock In {\em Proceedings of the IEEE/CVF Conference on Computer Vision and
  Pattern Recognition}, pages 16727--16738, 2023.

\bibitem{ye2022gifs}
Jianglong Ye, Yuntao Chen, Naiyan Wang, and Xiaolong Wang.
\newblock Gifs: Neural implicit function for general shape representation.
\newblock In {\em Proceedings of the IEEE/CVF Conference on Computer Vision and
  Pattern Recognition}, pages 12829--12839, 2022.

\bibitem{zhao2021learning}
Fang Zhao, Wenhao Wang, Shengcai Liao, and Ling Shao.
\newblock Learning anchored unsigned distance functions with gradient direction
  alignment for single-view garment reconstruction.
\newblock In {\em Proceedings of the IEEE/CVF International Conference on
  Computer Vision}, pages 12674--12683, 2021.

\bibitem{zhou2023learning}
Junsheng Zhou, Baorui Ma, Shujuan Li, Yu-Shen Liu, and Zhizhong Han.
\newblock Learning a more continuous zero level set in unsigned distance fields
  through level set projection.
\newblock In {\em Proceedings of the IEEE/CVF international conference on
  computer vision}, pages 3181--3192, 2023.

\bibitem{zhou2022learning}
Junsheng Zhou, Baorui Ma, Yu-Shen Liu, Yi Fang, and Zhizhong Han.
\newblock Learning consistency-aware unsigned distance functions progressively
  from raw point clouds.
\newblock {\em Advances in Neural Information Processing Systems},
  35:16481--16494, 2022.

\bibitem{zhou2022cap-udf}
Junsheng Zhou, Baorui Ma, Yu-Shen Liu, Yi Fang, and Zhizhong Han.
\newblock Learning consistency-aware unsigned distance functions progressively
  from raw point clouds.
\newblock In {\em Advances in Neural Information Processing Systems (NeurIPS)},
  2022.

\bibitem{zhou2018open3d}
Qian-Yi Zhou, Jaesik Park, and Vladlen Koltun.
\newblock {Open3D}: {A} modern library for {3D} data processing.
\newblock {\em arXiv:1801.09847}, 2018.

\bibitem{zhu2020deepfashion}
Heming Zhu, Yu Cao, Hang Jin, Weikai Chen, Dong Du, Zhangye Wang, Shuguang Cui,
  and Xiaoguang Han.
\newblock Deep fashion3d: {A} dataset and benchmark for 3d garment
  reconstruction from single images.
\newblock {\em CoRR}, 2020.

\end{thebibliography}


\begin{thebibliography}{10}\itemsep=-1pt

\bibitem{bhatnagar2019mgn}
Bharat~Lal Bhatnagar, Garvita Tiwari, Christian Theobalt, and Gerard Pons-Moll.
\newblock Multi-garment net: Learning to dress 3d people from images.
\newblock In {\em {IEEE} International Conference on Computer Vision ({ICCV})}.
  {IEEE}, oct 2019.

\bibitem{blinn1977models}
James~F. Blinn.
\newblock Models of light reflection for computer synthesized pictures.
\newblock page 192–198, 1977.

\bibitem{shapenet2015}
Angel~X Chang, Thomas Funkhouser, Leonidas Guibas, Pat Hanrahan, Qixing Huang,
  Zimo Li, Silvio Savarese, Manolis Savva, Shuran Song, Hao Su, et~al.
\newblock Shapenet: An information-rich 3d model repository.
\newblock {\em arXiv preprint arXiv:1512.03012}, 2015.

\bibitem{guillard2022meshudf}
Benoit Guillard, Federico Stella, and Pascal Fua.
\newblock Meshudf: Fast and differentiable meshing of unsigned distance field
  networks, 2022.

\bibitem{park2019deepsdf}
Jeong~Joon Park, Peter Florence, Julian Straub, Richard Newcombe, and Steven
  Lovegrove.
\newblock Deepsdf: Learning continuous signed distance functions for shape
  representation.
\newblock In {\em Proceedings of the IEEE/CVF conference on computer vision and
  pattern recognition}, pages 165--174, 2019.

\bibitem{sitzmann2020implicit}
Vincent Sitzmann, Julien Martel, Alexander Bergman, David Lindell, and Gordon
  Wetzstein.
\newblock Implicit neural representations with periodic activation functions.
\newblock {\em Advances in neural information processing systems},
  33:7462--7473, 2020.

\bibitem{ward1992anisotropic}
Gregory~J. Ward.
\newblock Measuring and modeling anisotropic reflection.
\newblock 26(2):265–272, 1992.

\bibitem{xu2019disn}
Qiangeng Xu, Weiyue Wang, Duygu Ceylan, Radomir Mech, and Ulrich Neumann.
\newblock Disn: Deep implicit surface network for high-quality single-view 3d
  reconstruction.
\newblock {\em CoRR}, abs/1905.10711, 2019.

\bibitem{zhou2022cap-udf}
Junsheng Zhou, Baorui Ma, Yu-Shen Liu, Yi Fang, and Zhizhong Han.
\newblock Learning consistency-aware unsigned distance functions progressively
  from raw point clouds.
\newblock In {\em Advances in Neural Information Processing Systems (NeurIPS)},
  2022.

\bibitem{zhu2020deepfashion}
Heming Zhu, Yu Cao, Hang Jin, Weikai Chen, Dong Du, Zhangye Wang, Shuguang Cui,
  and Xiaoguang Han.
\newblock Deep fashion3d: {A} dataset and benchmark for 3d garment
  reconstruction from single images.
\newblock {\em CoRR}, 2020.

\end{thebibliography}
}

\end{document}


\makeatletter

\makeatother

\newcommand{\cm}{\checkmark}

\newcommand{\newterm}[1]{\emph{#1}}
\newcommand{\TODO}[1]{$<$\textcolor{red}{#1}$>$}
\newcommand{\REV}[1]{\textcolor{blue}{#1}}
\newcommand{\REVNO}[1]{\st{#1}}
\newcommand{\final}[1]{#1}
\newcommand{\migue}[1]{[Migue]\textcolor{red}{#1}}
\newcommand{\emma}[1]{[Emma]\textcolor{blue}{#1}}
\newcommand{\viv}[1]{[Viv]\textcolor{cyan}{#1}}

\title{Supplementary material \\ DUDF: Differentiable Unsigned Distance Fields with Hyperbolic Scaling}

\author{Miguel Fainstein$^{\text{1},\text{2}}$\\
{\tt\small miguelon.f98@gmail.com}
\and
Viviana Siless$^{\text{1}}$\\
{\tt\small viviana.siless@utdt.edu}
\and
Emmanuel Iarussi$^{\text{1},\text{3}}$\\
{\tt\small emmanuel.iarussi@utdt.edu}
\and
$^{\text{1}}$Universidad Torcuato Di Tella, $^{\text{2}}$FCEyN Universidad de Buenos Aires, $^{\text{3}}$CONICET
} 
\maketitle

\section{Differentiability analysis}

\label{sec:diff_an}
Given a smooth surface $\mathcal{S}$, we want to study the differentiability of function: 
\begin{equation}
    \label{eq:ts}
    t_\mathcal{S}(\mathbf{x})=d_\mathcal{S}(\mathbf{x}) \tanh( \alpha d_\mathcal{S}(\mathbf{x}) ).
\end{equation}
Notice that function $d_\mathcal{S}$ is differentiable almost everywhere outside the surface, hence $t_\mathcal{S}$ is trivially differentiable at such points and its gradient is given by:
\begin{equation}
        \nabla t_\mathcal{S}(\mathbf{x}) = \nabla d_\mathcal{S}(\mathbf{x}) \phi(\mathbf{x}),
\end{equation}
with \small{$\phi(\mathbf{x}) = \tanh(\alpha d_\mathcal{S}(\mathbf{x}))+ \alpha d_\mathcal{S}(\mathbf{x})(1 - \tanh^2(\alpha d_\mathcal{S}(\mathbf{x})))$}.
However, at the isosurface $d_\mathcal{S}$ is not differentiable. To show $t_\mathcal{S}$ differentiable at $\mathbf{s} \in \mathcal{S}$, we consider the partial derivatives expressed by the limit:
\begin{equation}
    \lim_{h\to0} \dfrac{t_\mathcal{S}(\mathbf{s} + \mathbf{e}_i h) - t_\mathcal{S}(\mathbf{s})}{h} = \lim_{h\to0} \dfrac{t_\mathcal{S}(\mathbf{s} + \mathbf{e}_i h)}{h},
\end{equation}
where $\mathbf{e}_i$ is the $i$-th canonical vector. 
It follows that  $d_\mathcal{S}(\mathbf{s} + \mathbf{e}_i h) \leq \left | h \right |$. Considering that function $\tanh$ is monotonically increasing, then:

\begin{equation}
    \left | \dfrac{t_\mathcal{S}(\mathbf{s} + \mathbf{e}_i h)}{h} \right | \le \left | \dfrac{\left | h \right | \tanh( \alpha \left | h \right | )}{h} \right | = \left | \tanh( \alpha \left | h \right | ) \right | \xrightarrow[]{h\to0} 0.
\end{equation}

Hence, partial derivatives at the isosurface are null. To finish the proof, we show that the hyperplane defined by the partial derivatives at $\mathbf{s}$ correctly approximates the function $t_\mathcal{S}$, which is satisfied if the following limit approaches 0: 
\begin{equation}
    \label{eq:hyperplane}
    \lim_{\mathbf{x}\to\mathbf{s}} \frac{\left | t_\mathcal{S}(\mathbf{x}) - t_\mathcal{S}(\mathbf{s}) - \nabla t_\mathcal{S}(\mathbf{s}) \cdot (\mathbf{x} - \mathbf{s}) \right |}{\left \| \mathbf{x} - \mathbf{s} \right \|} = \lim_{\mathbf{x}\to\mathbf{s}} \frac{\left | t_\mathcal{S}(\mathbf{x}) \right |}{\left \| \mathbf{x} - \mathbf{s} \right \|}.
\end{equation}
Similar to the partial derivative calculation, we bound this limit considering that the distance of every point to the surface must always be smaller or equal to the euclidean distance between a point and some surface point, that is $d_\mathcal{S}(\mathbf{x}) \leq \left \| \mathbf{x} - \mathbf{s} \right \|$ for every $\mathbf{s} \in \mathcal{S}$. Using this fact in Eq. \ref{eq:hyperplane}:

\small{
\begin{equation}
    \frac{\left | t_\mathcal{S}(\mathbf{x}) \right |}{\left \| \mathbf{x} - \mathbf{s} \right \|} \leq \frac{ \left \| \mathbf{x} - \mathbf{s} \right \| \left | \tanh(\alpha d_\mathcal{S}(\mathbf{x})) \right |}{\left \| \mathbf{x} - \mathbf{s} \right \|} = \left | \tanh(\alpha d_\mathcal{S}(\mathbf{x})) \right | \xrightarrow[]{\mathbf{x}\to\mathbf{s}} 0,
\end{equation}}
concluding that function $t_\mathcal{S}$ is differentiable at the isosurface.

\section{Gradient norm}
Based on the results from Sec. \ref{sec:diff_an}, at points outside the isosurface where $d_\mathcal{S}$ is differentiable, the unsigned distance gradient norm is unitary: $\left \| \nabla d_\mathcal{S}(\mathbf{x}) \right \|=1$. This is a standard property of distance functions.
Given that $\phi$ is always positive away from the isosurface, the norm of the gradient can be expressed as:
\begin{equation}
    \left \| \nabla t_\mathcal{S}(\mathbf{x}) \right \| = \left \| \nabla d_\mathcal{S}(\mathbf{x}) \right \| \left | \phi(\mathbf{x}) \right | = \phi(\mathbf{x}).
\end{equation}
As previously discussed, at the isosurface the gradient $\nabla t_\mathcal{S}(\mathbf{s}) = \mathbf{0}$, then its norm is null. 
These results explain the choice for the \textit{Eikonal} problem and \textit{Neumann} boundary condition expressed in the paper.

\section{Extended ablation and results}
As mentioned in the paper, we experimented with different values for parameter $\alpha$ (Tab. \ref{tab:alfa}).
As $\alpha$ gets larger, function $t_\mathcal{S}$ closely approximates $d_\mathcal{S}$, increasing the errors 
probably due to the non-differentiability at the isosurface. 
Smaller $\alpha$ values enlarge the quadratic strip near the isosurface, which is harder to supervise effectively and hinders the performance of MC.  

\setlength{\columnsep}{10pt}%
\setlength{\intextsep}{10pt}%
\begin{wraptable}{r}{3.5cm}
    \centering
    \resizebox{0.45\columnwidth}{!}{
    \setlength\tabcolsep{0.5pt}
    \begin{tabular} {  c c c c } 
    \toprule
    $\alpha$ & L1CD $\downarrow$ & L2CD $\downarrow$ & NC $\downarrow$ \\
    \hline
    $1\times 10^0$ & $10.7$ & $0.038$ & $0.094$ \\
    $1\times 10^1$ & $9.23$ & $\mathbf{0.025}$ & $0.033$ \\
    $1\times 10^2$ & $\mathbf{9.14}$ & $0.027$ & $\mathbf{0.020}$ \\
    $1\times 10^3$ & $9.52$ & $0.038$ & $0.023$ \\
    $1\times 10^4$ & $9.69$ & $0.029$ & $0.028$ \\
    \bottomrule
    \end{tabular}}
    \caption{L1, L2 Chamfer distance and Normal Consistency (NC) metrics for different values of parameter $\alpha$.}
    \label{tab:alfa}
\end{wraptable}

\begin{figure*}[ht]
  \centering
  \includegraphics[width=1.0\linewidth]{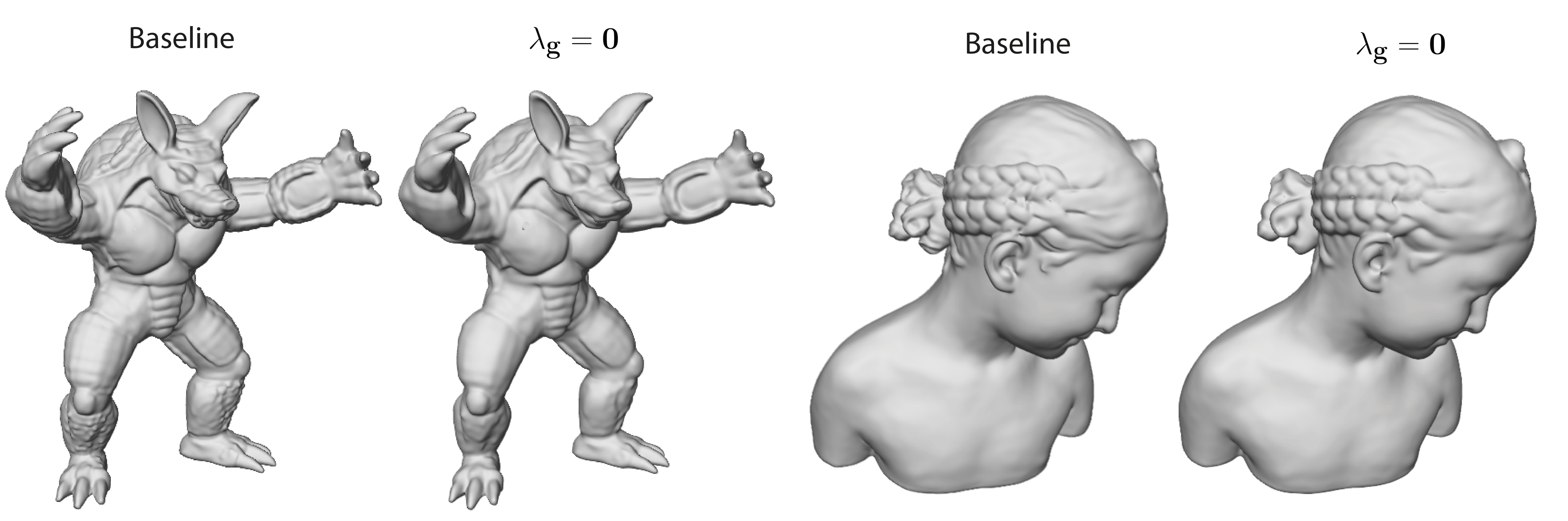}
   \caption{Ablation study. The first column shows the reconstruction of the network trained with the full loss. The second column shows the reconstruction without the maximum curvature field alignment term, resulting in less detail at high curvature regions such as Armadillo's mouth and Bimba's hair bun.}
   \label{fig:ablation}
\end{figure*}

\begin{figure*}[ht]
  \centering
  \includegraphics[width=1.0\linewidth]{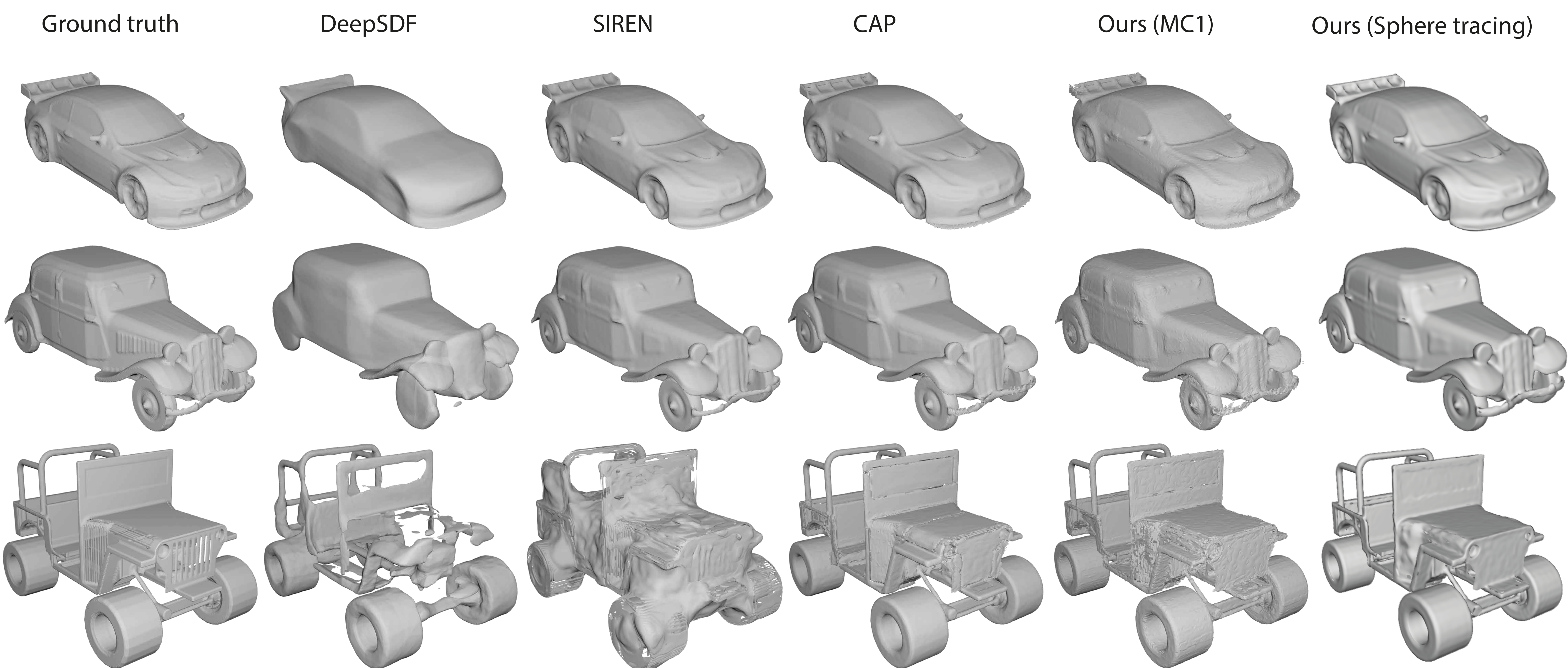}
   \caption{Reconstructions on closed \cite{xu2019disn} ShapeNet cars \cite{shapenet2015}
   for DeepSDF \cite{park2019deepsdf}, SIREN \cite{sitzmann2020implicit}, CAP-UDF \cite{zhou2022cap-udf} and DUDF. We additionally show sphere tracing renderings of our learned fields without the marching cubes reconstruction step.}
   \label{fig:closed}
\end{figure*}

\begin{figure*}[ht]
  \centering
  \includegraphics[width=1.0\linewidth]{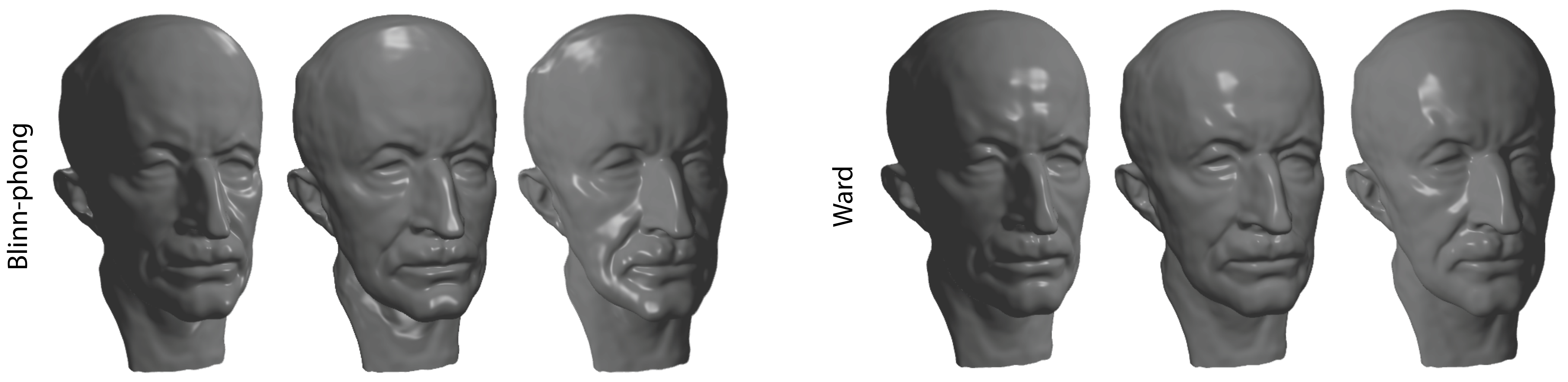}
   \caption{Sphere tracing renderings of an open surface (Max Planck model is open at the bottom) under three different light sources. We used surface normals for the Blinn-phong reflectance model \cite{blinn1977models}. For the Ward reflectance model \cite{ward1992anisotropic} we used both, 
   normals and principal curvature directions computed from DUDF's differentiable fields.}
   \label{fig:iluminacion}
\end{figure*}

\begin{figure*}[ht]
  \centering
  \includegraphics[width=1.0\linewidth]{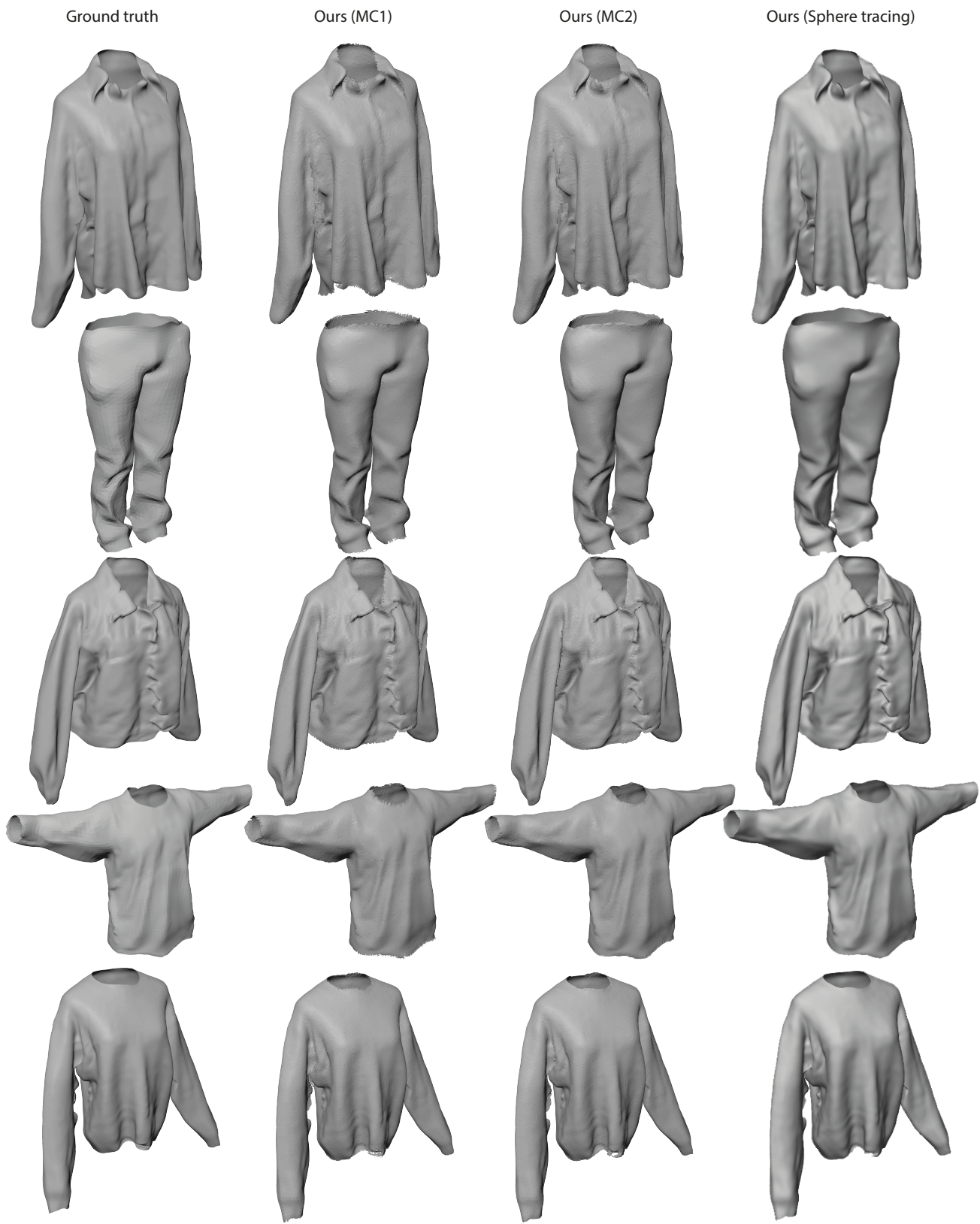}
   \caption{Reconstructions on  DeepFashion \cite{zhu2020deepfashion} and Multi-garment \cite{bhatnagar2019mgn} data sets for the three different reconstruction methods explored in our work: MC1\cite{zhou2022cap-udf}, MC2\cite{guillard2022meshudf}, and sphere tracing.}
   \label{fig:methods}
\end{figure*}

We also show additional qualitative comparisons for three different experiments showcased in the paper. In Fig. \ref{fig:ablation} we demonstrate the effect of the maximum curvature field alignment loss on two reconstructed examples. 
In Fig. \ref{fig:closed} we compare closed surface representations trained on the ShapeNet Cars data set\cite{shapenet2015} pre-processed by \cite{xu2019disn} to be closed. 
In Fig. \ref{fig:iluminacion} we show how our field's normal directions and principal curvatures are useful in the context of tracing rendering algorithms and light reflection models.
Finally, in Fig. \ref{fig:methods} we further compare the different reconstruction algorithms used throughout this work. 
All reconstructions were performed with the parameters reported in the paper (architecture, parameter count, loss weights, and training scheme).

\newpage

\newpage

\clearpage

{\small
\bibliographystyle{ieee_fullname}
\bibliography{egbib}
}